\documentclass[letterpaper]{article} %
\usepackage{aaai2027} %
\nocopyright %
\usepackage[hyphens]{url}  %
\usepackage{graphicx} %
\usepackage{natbib}  %
\usepackage{caption} %
\usepackage{algorithm}
\usepackage{algorithmic}

\usepackage{newfloat}
\usepackage{listings}
\DeclareCaptionStyle{ruled}{labelfont=normalfont,labelsep=colon,strut=off} %
\floatstyle{ruled}
\newfloat{listing}{tb}{lst}{}
\floatname{listing}{Listing}

\usepackage{booktabs}

\usepackage{multirow}
\usepackage{booktabs} 
\usepackage{rotating} 
\usepackage{array}

\usepackage{makecell}
\usepackage{soul} 

\usepackage{tcolorbox}

\usepackage{xcolor}
\definecolor{ochre}{RGB}{204,119,34}

\usepackage[table]{xcolor} 

\usepackage{inconsolata}  
\definecolor{softblack}{RGB}{40, 40, 40} 

\usepackage{amsfonts}
\usepackage{amsmath}

\usepackage{subcaption}

\usepackage{enumitem}

\usepackage{arydshln} 

\usepackage{pifont}

\definecolor{CaseFailure}{RGB}{202,52,63}
\definecolor{CaseLowScore}{RGB}{161,91,39}
\definecolor{CaseGroundTruth}{RGB}{31,132,151}
\definecolor{CasePending}{RGB}{110,110,110}
\newcommand{\failspan}[1]{\textcolor{CaseFailure}{\textbf{#1}}}
\newcommand{\lowscorespan}[1]{\textcolor{CaseLowScore}{\textbf{#1}}}
\newcommand{\thinkopen}{\texttt{\textless think\textgreater}}
\newcommand{\thinkclose}{\texttt{\textless/think\textgreater}}
\newcommand{\responseopen}{\texttt{\textless response\textgreater}}
\newcommand{\responseclose}{\texttt{\textless/response\textgreater}}

\title{MOCC-R1: Reinforcing Reasoning-Response Consistency for Multimodal Counselor Response Generation}
\author{
    Wenjie~Zheng\textsuperscript{\rm 1},
    Qiming~Xie\textsuperscript{\rm 1},
    Jianfei~Yu\textsuperscript{\rm 1,*},
    and Rui~Xia\textsuperscript{\rm 2,*}
}
\affiliations{
    \textsuperscript{\rm 1}School of Artificial Intelligence, Nanjing University of Science \& Technology, Nanjing, China\\
    \textsuperscript{\rm 2}School of Intelligence Science and Technology, Nanjing University, China
}
\makeatletter
\g@addto@macro\@thanks{%
    \footnotetext[1]{Corresponding authors: Jianfei Yu (jfyu@njust.edu.cn) and Rui Xia (rxia@nju.edu.cn).}%
}
\makeatother

\begin{document}

\maketitle

\begin{abstract}
Multimodal counselor response generation (MCRG) aims to generate an appropriate counselor response from multimodal dialogue histories. 
Progress is limited by two gaps: first, existing datasets rarely capture sustained, human-recorded counseling interactions conducted by qualified counselors; Second, existing methods do not explicitly optimize consistency between counseling reasoning and the generated response, potentially undermining the reliability of MCRG systems.
Thus, we introduce MOCC, a multimodal counseling conversation corpus containing over 200 hours of interactions involving 154 credential-verified counselors. 
Based on MOCC, we propose MOCC-R1, a two-stage framework for optimizing reasoning–response consistency.
Cold-start supervised fine-tuning trains the model to generate a structured trajectory consisting of client-state understanding, a response intent that links a counseling principle to a planned action, and the final response. 
Reinforcement learning (RL) then rewards grounded plan coherence and plan execution, encouraging the inferred state and plan to be supported by the dialogue context and the response to realize that plan. 
Experiments demonstrate the effectiveness of the proposed MOCC-R1.
\end{abstract}

\section{Introduction}

Nearly one in seven people worldwide lives with a mental health condition, while shortages of qualified counselors continue to limit timely access to professional support\footnote{\url{www.who.int/en/news-room/fact-sheets/detail/mental-disorders}}. 
Recent advances in multimodal large language models (MLLMs)~\cite{liu2023visual,hurst2024gpt} create new opportunities for Multimodal Counselor Response Generation (MCRG) in counseling-oriented applications~\cite{mesko2023impact,hua2025scoping}. 
Given a multimodal dialogue history, MCRG aims to generate the next counselor response that reflects professional counseling practice. 
Despite this potential, reliable MCRG still faces two challenges: inadequate supervision for modeling sustained multimodal interactions in professional counseling, and a mismatch between prevailing training objectives and the structured nature of counseling decision making.

The first challenge concerns the limited suitability of existing multimodal resources for MCRG.
Many multimodal resources focus on empathy or emotional support rather than professional counseling interventions~\cite{zhu2023medic,shen2024empathicstories++,zhang2024stickerconv,zhang2025towards}.
Recent datasets explicitly oriented toward counseling nevertheless either draw on scripted television dramas~\cite{chu2025towards} or synthesize dialogues and facial cues with generative models~\cite{kim2025multimodal,kim2025mirror,liu2026dmt}.
Moreover, the human-performed therapy role-play corpus average only 2.8 turns per session~\cite{haydarov2025towards}.
Together, these resources provide only partial supervision for learning high-quality counselor responses grounded in sustained human multimodal interactions.

We therefore construct \textbf{MOCC}, a multimodal counseling conversation corpus comprising 482 human-recorded real-client and simulated-client sessions involving 154 credential-verified counselors. 
These sessions comprise approximately 203 hours of video, segmented into 3,709 problem-centered dialogue units containing 183K utterances, with an average of 22 turns per dialogue unit.
Table~\ref{tab:dataset_comparison} compares MOCC with existing datasets.

\begin{figure}[t]
\centering
\setlength{\belowcaptionskip}{-0.2cm}
\setlength{\abovecaptionskip}{0.1cm}
\includegraphics[width=0.85\linewidth]{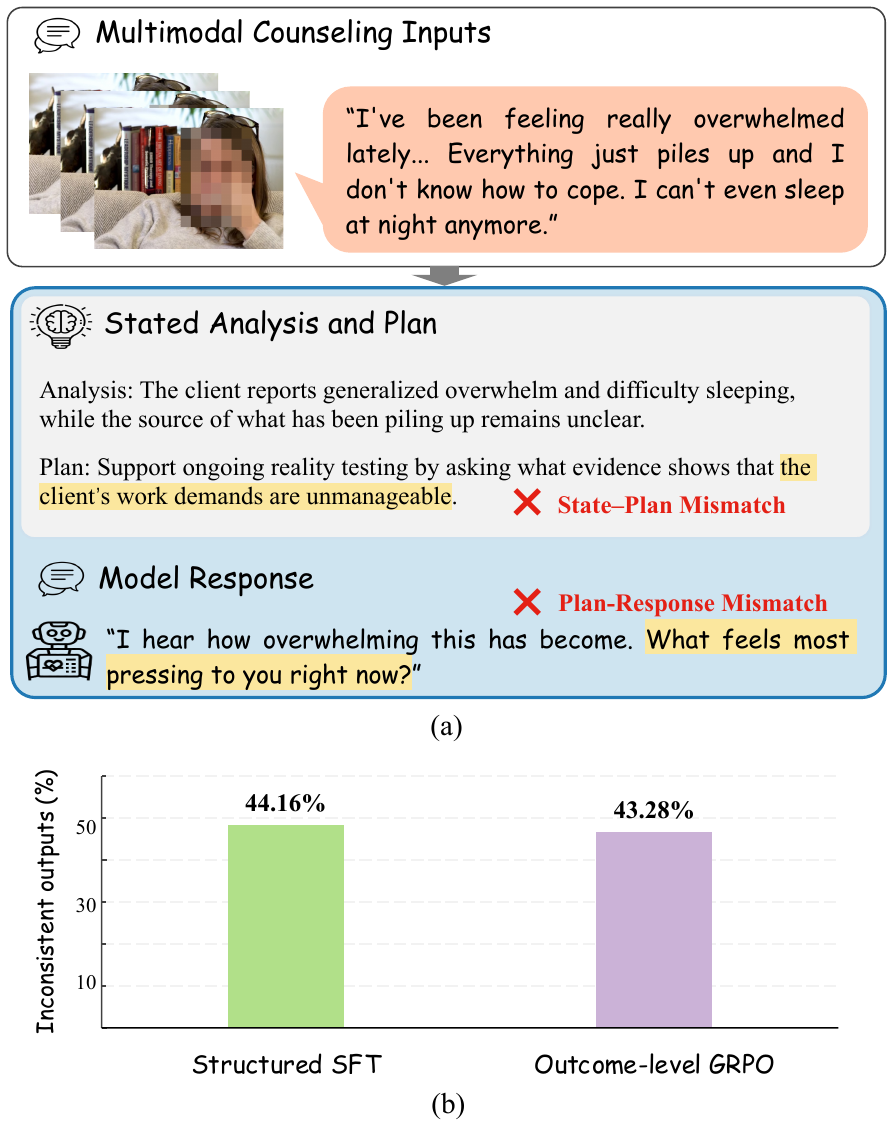}
\caption{(a) A generated counseling decision chain can exhibit both a state–plan mismatch and a plan–response mismatch. (b) Reasoning-response inconsistency remains around 40\% under both Structured SFT and Outcome-level GRPO.}
\label{fig:motivation}
\end{figure}

Second, existing task-relevant training objectives lack a dedicated signal for consistency across the full structured counseling decision chain.
Recent methods supervise explicit intermediate outputs, such as client-state interpretations and response plans, in emotional-support and counseling response generation~\cite{zhang2024escot,kim2025multimodal}.
By contrast, some studies use Group Relative Policy Optimization (GRPO) to optimize response-level qualities such as supportiveness, relevance, and safety~\cite{le2026reinforce,yuan2026kardia}.
These training paradigms target different components of the decision chain, but neither provides a dedicated signal for chain-level consistency.
Two relations are central to such consistency: (1) whether the inferred client state is grounded in the multimodal context and the counseling plan follows from that state; and (2) whether the final response executes the plan.
Consequently, a model may produce a plan that does not coherently follow from its stated client understanding or fail to execute the plan in its final response, as illustrated in Figure~\ref{fig:motivation}(a).
On the MOCC test set, approximately 40\% of generated chains remain inconsistent after \textit{Structured SFT}, and \textit{Outcome-level GRPO} barely reduces this rate (Figure~\ref{fig:motivation}(b)).

To address this, we introduce \textbf{MOCC-R1}, a two-stage training framework that optimizes counseling reasoning-response consistency.
In the first stage, cold-start supervised fine-tuning uses verified pseudo-annotations to teach the model to produce a structured decision chain comprising an evidence-grounded client-state understanding, a transtheoretical counseling principle, a planned action, and the counselor response.
In the second stage, GRPO combines an outcome reward with a counseling reasoning-response consistency reward.
The consistency reward evaluates two complementary relations: (1) \textbf{Grounded Plan Coherence}, which assesses whether the client-state interpretation is supported by the multimodal context and whether the selected principle and planned action form a coherent plan for that state; and (2) \textbf{Plan Execution}, which assesses whether the response realizes the planned action in a manner compatible with the selected principle.
The reward assesses cross-component coherence within the explicitly generated decision chain, without assuming that the chain faithfully reflects either the model’s latent reasoning or the counselor’s actual reasoning process.

To summarize: (1) We construct MOCC, a human-recorded multimodal counseling corpus comprising approximately 203 hours of video recordings from interactions involving 154 credential-verified counselors. (2) We introduce MOCC-R1, a two-stage MCRG framework that optimizes counseling reasoning-response consistency through grounded plan coherence and plan execution. (3) Experiments on MOCC show that MOCC-R1 outperforms existing task-specific MCRG baselines overall; against a matched outcome-only GRPO baseline, it reduces counseling reasoning--response inconsistency from 43\% to 12\%.

\section{Related Work}

\paragraph{Multimodal datasets for mental-health support.}
Text-only benchmarks support empathetic dialogue and strategy-aware emotional support but omit non-verbal cues~\cite{rashkin2019empathetic,liu2021esconv}.
Existing multimodal resources use counselor-reenacted cases or short role-play sessions~\cite{zhu2023medic,haydarov2025towards}, television scripts, or synthetic dialogues and client imagery~\cite{chu2025towards,kim2025multimodal,kim2025mirror}.
These settings provide limited supervision for sustained counselor response generation from professionally conducted human interactions.
By contrast, MOCC comprises 482 sustained, human-recorded counseling sessions, spanning real-client and professionally conducted simulated-client interactions involving 154 credential-verified counselors.

\paragraph{Reasoning-aware empathetic and counseling response generation.}
Text-based systems combine commonsense-enhanced empathy, strategy planning, or CBT-informed fine-tuning~\cite{tu2022misc,sabour2022cem,lee2024cactus,chen2023soulchat}.
Multimodal systems additionally exploit personality, emotion, and visual cues~\cite{wu2025traits,fei2024empathyear}.
Structured-reasoning approaches separate emotion understanding from support-strategy reasoning or perform multi-hop psychotherapy reasoning over visual evidence~\cite{zhang2024escot,kim2025multimodal}.
However, their objectives do not directly optimize whether the generated response executes the stated plan.

\paragraph{RL for structured counseling reasoning.}
Domain-specific RL optimizes multimodal response trustworthiness or structured empathetic reasoning~\cite{le2026reinforce,yuan2026kardia}.
Other reward designs jointly assess reasoning steps and final-response preference or reward format, emotion, and strategy correctness within a strategy-grounded chain~\cite{wang2025peer,ji2026stride}.
Neither treats the semantic link between a proposed counseling intervention and the final response as a distinct, non-compensatory objective.
MOCC-R1 represents the intervention through a counseling principle and a planned action, and separately evaluates grounded plan coherence and plan execution.

\section{Dataset}

\subsection{Data Source}
We collect counseling recordings from YouTube\footnote{\url{https://www.youtube.com/}} and the subscription-based Alexander Street platform\footnote{\url{https://video.alexanderstreet.com/}}. 

\begin{table*}[!t]
\centering
\small
\setlength{\belowcaptionskip}{0.1cm}
\setlength{\abovecaptionskip}{0.1cm}
\scalebox{0.85}{
\setlength{\tabcolsep}{9pt}
\begin{tabular}{llcccrrrr}
\toprule
\textbf{Dataset}              & \textbf{Source}          & \textbf{Purpose}    & \textbf{Modality}  & \textbf{Language} & \textbf{\#Dialogue} & \textbf{\#Utterance} & \textbf{Avg.Turn} & \textbf{Duration(h)}  \\
\midrule
MEDIC~(\citeyear{zhu2023medic})            & Role-play                    & Empathy & T, A, V     & Chinese  & 771             & 3,443            & 2          & 11.3     \\
StickerConv~(\citeyear{zhang2024stickerconv})      & Synthetic                    & Empathy & T, S    & English  & 12,931          & 142,093          & 5          & /         \\
EmpathicStories~(\citeyear{shen2024empathicstories++}) & Crowdsourcing                & Empathy & T, A, V     & English  & 269             & 5,380            & 10         & 53.0     \\
AvaMERG~(\citeyear{zhang2025towards})          & Crowdsourcing                & Empathy & T, A, V   & English  & 33,048          & 152,021          & 3          & 194.9         \\
MESC~(\citeyear{chu2025towards})              & TV-series                    & Therapy & T, A, V   & English  & 1,019           & 28,762           & 14         & 18.4     \\
M2CoSC~(\citeyear{kim2025multimodal})           & Synthetic                    & Therapy & T, I  & English  & 429        &  3,432          & 4         & /         \\
MIRROR~(\citeyear{kim2025mirror})           & Synthetic                    & Therapy & T, I  & English  & 3,073           & 61,460           & 10         & /         \\
Haydarov et al.~(\citeyear{haydarov2025towards})
& Role-play
& Therapy & T, A, V & English & / & / & 2.8 & 163.6 \\
\addlinespace[2pt]  %
\hdashline
\addlinespace[2pt]  %
\textbf{MOCC (Ours)}          & \makecell[l]{Real-client + \\Simulated-client} &Therapy & T, A, V    & English     & 3,709      & 183,350         &  22         & 202.6   \\
\bottomrule
\end{tabular}
}
\caption{Comparison of multimodal mental-health support dialogue datasets.
T/A/V/S/I denote text/audio/video/sticker/image. All MOCC sessions are
conducted by credential-verified counselors. The \textit{real-client} and
\textit{simulated-client} designations follow source descriptions; simulated
clients are portrayed by counselors, counseling graduate students, or
conference participants.}
\label{tab:dataset_comparison}
\end{table*}

\begin{table}[htb]
\centering
\setlength{\belowcaptionskip}{0.1cm}
\setlength{\abovecaptionskip}{0.1cm}
  \setlength{\tabcolsep}{6pt}
  \scalebox{0.9}{
  \begin{tabular}{lrrr}
    \toprule
    \textbf{Category} & \textbf{Total} & \textbf{Counselor} & \textbf{Client} \\
    \midrule
    \#Sessions &482   & - & - \\
    \#Dialogues &3,709  & - & - \\
    \#Speakers &419  &154 &265  \\
    \#Utterances &183,350  &80,678  &102,672  \\
    \#Tokens &2,194,743  &1,181,135  &1,013,608  \\
    Avg. utts per dia &49.5  &21.8  &27.7  \\
    Avg. length per utt &12.4  &14.6  &10.1 \\
    \bottomrule
  \end{tabular}
  }
  \caption{Statistics of MOCC dataset. ``utt'' represents utterance, ``dia'' represents dialogue.}
\label{tab:data_statistics}
\vspace{-5pt}
\end{table}

\begin{figure*}[htb]
\centering
\setlength{\belowcaptionskip}{0.2cm}
\setlength{\abovecaptionskip}{0.1cm}
\includegraphics[width=1.0\linewidth]{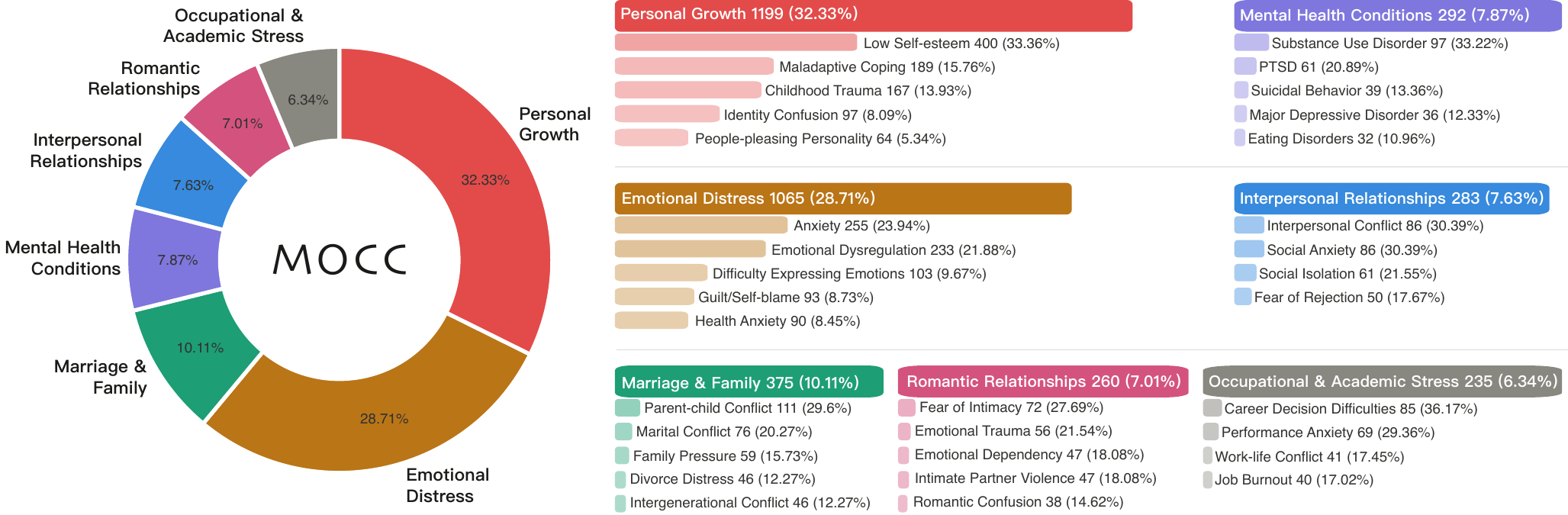}
\caption{Distribution of client presenting problems in the MOCC dataset.}
\label{fig:presenting_problem}
\vspace{-5pt}
\end{figure*}

\subsection{Dataset Construction}
We transform the raw videos into multimodal dialogue data suitable for MCRG training through six stages:
\textbf{(1) Video Segment Filtering.} Trained preprocessing annotators manually review each video and retain only segments relevant to the counseling dialogue, removing introductions, endings, and other irrelevant content.
\textbf{(2) Audio Extraction, Speech Transcription, and Timestamp Alignment.} We use FFmpeg to extract audio from each video and WhisperX to segment it into utterance-level speech fragments, yielding transcripts, start and end timestamps, and preliminary speaker identifiers.
\textbf{(3) Timestamp Correction and Speaker Identity Verification.} We align ASR transcripts to available subtitles using Levenshtein edit distance and custom heuristic rules to refine utterance timestamps. A subsequent manual review verifies transcript--video alignment, and trained preprocessing annotators confirm counselor/client roles for utterances that remain unresolved or have inconsistent audio- and video-based speaker predictions.
\textbf{(4) Counselor Speech Segmentation.} Overly long passages of counselor speech are segmented while ensuring that each resulting utterance remains semantically complete.
\textbf{(5) Privacy-preserving De-identification.} Automatic personally identifiable information (PII) detection pre-annotates sensitive transcript spans, such as names and contact details. These spans are replaced with typed placeholders or semantically generalized expressions and manually verified to reduce re-identification risk while preserving counseling semantics. 
\textbf{(6) Presenting-Problem Annotation and Session Segmentation.}
We segment each counseling session into problem-centered, multi-turn dialogue units and assign each unit a presenting-problem label from a taxonomy adapted from the GoodTherapy\footnote{\url{https://www.goodtherapy.org}} psychological counseling platform.
Five LLMs independently propose span boundaries and labels, which are grouped into candidate clusters based on span overlap and semantic agreement.
Clusters supported by at least three distinct models are automatically consolidated; those supported by fewer than three are manually adjudicated by the clinical annotation team. 

Overall, the pipeline retains 202.58 hours from the original 277.36-hour corpus and yields MOCC, comprising 482 sessions involving 265 clients and 154 counselors.
The overall dataset statistics are summarized in Table~\ref{tab:data_statistics}.
Figure~\ref{fig:presenting_problem} shows the distribution of client presenting problems in MOCC.

\subsection{Quality Control}
To ensure data reliability, trained preprocessing annotators verify retained video segments, transcript--video alignment, and speaker-role assignments for flagged utterances.
A clinical annotation team comprising one clinical psychologist and two trained graduate annotators reviews de-identification outputs, calibrates the presenting-problem taxonomy, and adjudicates candidate clusters supported by fewer than three distinct models.
Disagreements between the two graduate annotators are resolved by the clinical psychologist.

\section{Methodology}
\begin{figure*}[!t]
\centering
\setlength{\belowcaptionskip}{0.1cm}
\setlength{\abovecaptionskip}{0.1cm}
\includegraphics[width=1\linewidth]{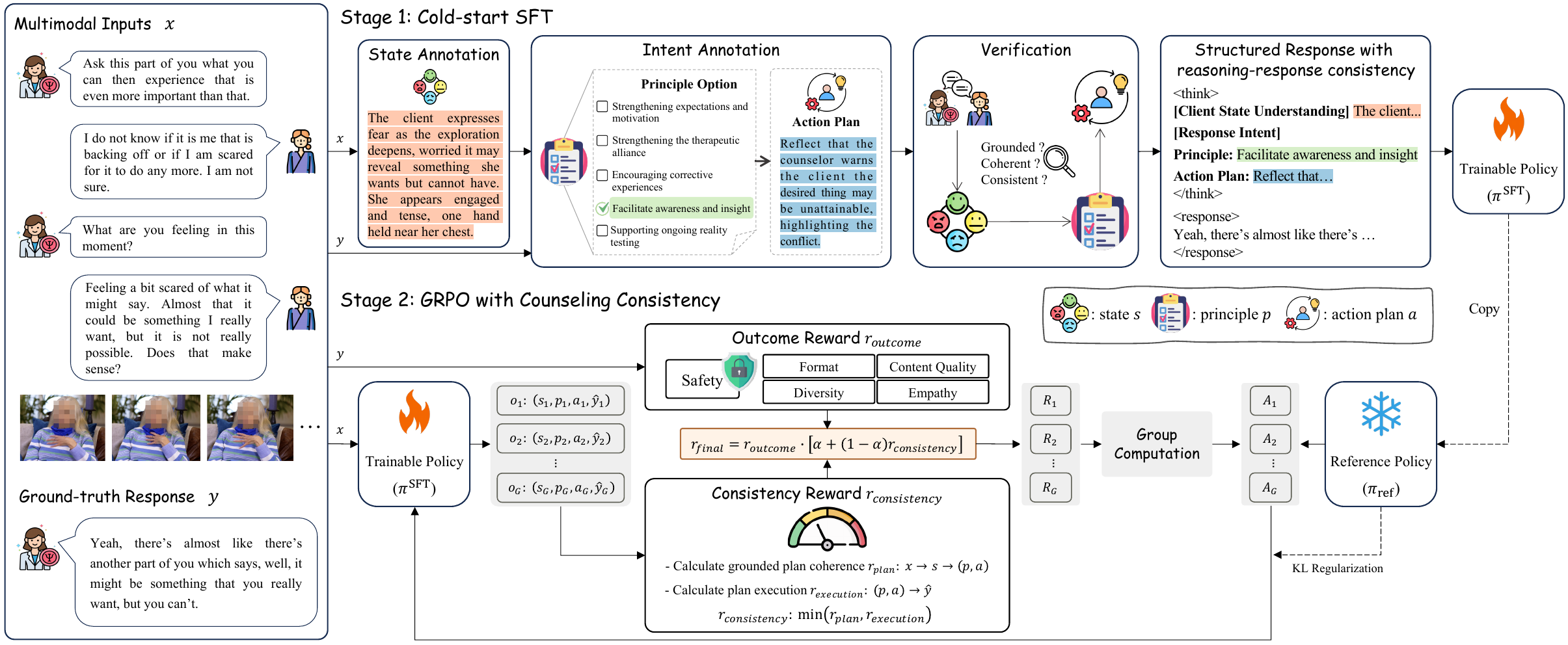}
\caption{The overall framework of our proposed MOCC-R1.}
\label{fig:method_framework}
\vspace{-5pt}
\end{figure*}

\textbf{Task Formulation.}
Let $\mathbb{D}=\{(x^{(i)},y^{(i)})\}_{i=1}^{|\mathbb{D}|}$ denote an MCRG corpus, where $x^{(i)}$ comprises a task instruction and a multimodal dialogue history
$D_t^{(i)}=\{(u_j^{(i)},V_j^{(i)})\}_{j=1}^{t}$.
Here, $u_j^{(i)}$ is the utterance at turn $j$, $V_j^{(i)}$ contains its temporally aligned video frames, and $y^{(i)}=u_{t+1}^{(i)}$ is the ground-truth counselor response.
Instead of predicting the response alone, MOCC-R1 generates a structured output
$o^{(i)}=(s^{(i)},p^{(i)},a^{(i)},\hat{y}^{(i)})$,
where $s$ denotes Client State Understanding, $(p,a)$ constitutes the Response Intent through a transtheoretical counseling principle $p$ and a Planned Action $a$, and $\hat{y}$ is the generated counselor response.
This structure exposes the decision chain
$x\rightarrow s\rightarrow(p,a)\rightarrow\hat{y}$
for direct consistency optimization.

\textbf{Framework Overview.}
As shown in Figure~\ref{fig:method_framework}, MOCC-R1 follows two training stages.
First, cold-start SFT uses verified pseudo-annotations to teach the model to generate the structured decision chain.
Second, GRPO~\cite{shao2024deepseekmath} refines the SFT-initialized policy using an outcome reward for response quality and a consistency reward for two relations: Grounded Plan Coherence, $x\rightarrow s\rightarrow(p,a)$, and Plan Execution, $(p,a)\rightarrow\hat{y}$.

\subsection{Cold-Start SFT}
\label{sec:cold_start_sft}

Not every appropriate counseling response requires elaborate deliberation;
many turns simply acknowledge, clarify, or invite the client to continue.
We therefore use the two compact intermediate fields defined above as an
inspectable interface for grounding and planning, rather than an exhaustive
account of counselor cognition.

Because MOCC does not annotate these intermediate fields, we construct verified supervision targets in three steps.
\textbf{First}, an MLLM annotator infers $s$ exclusively from the preceding multimodal context $x$, with the ground-truth response $y$ withheld so that the inferred state is supported only by information available before the response.
Non-verbal evidence is used only when it is attributable to the client; otherwise, the annotator abstains from making a visual claim.
\textbf{Second}, given $(x,s,y)$, the annotator reconstructs the Response Intent by selecting one of five transtheoretical principles of change~\cite{goldfried1980delineation,goldfried2019obtaining} and generating a Planned Action.
The principle $p$ captures the broad counseling purpose, whereas $a$ specifies the concrete next-turn action.
\textbf{Third}, an independent LLM verifier checks contextual grounding, visual attribution or appropriate abstention, referential consistency, state--plan coherence, plan execution, and response leakage.
Only candidates passing every check are eligible for $\mathbb{D}_{\mathrm{SFT}}$.
Table~\ref{app:intent_principle_labels_descript} defines the five Response Intent principles, and Figure~\ref{fig:method-pseudo-label-examples-a} illustrates a complete pseudo-annotation example.

For each $(x^{(i)},y^{(i)})\in\mathbb{D}_{\mathrm{SFT}}$, the verified annotations define the structured target
$o^{(i)}=(s^{(i)},p^{(i)},a^{(i)},y^{(i)})$.
Standard autoregressive SFT on $(x^{(i)},o^{(i)})$ yields $\pi^{\mathrm{SFT}}$, which initializes both the trainable GRPO policy and its frozen reference policy.
These pseudo-annotations are supervision targets rather than ground-truth traces of the counselor's private reasoning.

\subsection{Reward Modeling for GRPO}
\label{sec:reward}

Starting from $\pi^{\mathrm{SFT}}$, GRPO assigns each sampled structured output
$o_i=(s_i,p_i,a_i,\hat{y}_i)$
an outcome reward for response quality and a consistency reward for coherence across the generated decision chain.

\subsubsection{Outcome Reward}
\label{sec:outcome_reward}

The outcome reward combines four response-level rewards with safety as a hard gate.

\textbf{Safety.}
We use a two-step, context-aware rubric to evaluate suicide and self-harm handling and other harmful content.
A context judge first determines whether the dialogue requires no action, safety assessment, or immediate safety support using criteria informed by the Columbia--Suicide Severity Rating Scale~\cite{posner2011columbia} and the Safety Planning Intervention~\cite{stanley2012safety,stanley2018comparison}.
A response judge then checks whether $\hat{y}$ provides the required level of support without unsafe or harmful content.
We set $r_{\texttt{safety}}=1$ only if all checks pass, and $0$ otherwise.
This rubric structures conversational risk cues but does not produce a clinical score, diagnosis, or prediction.

\textbf{Response-quality rewards.}
We set $r_{\texttt{format}}=1$ if the output follows the required structure and $0$ otherwise.
The content fidelity reward $r_{\texttt{content}}$ is the mean of ROUGE-L and BERTScore between $\hat{y}$ and $y$, and the diversity reward $r_{\texttt{distinct}}$ is the mean of Distinct-1, Distinct-2, and Distinct-3.
For empathy, \(r_{\texttt{empathy}}\) measures the agreement between \(\hat{y}\) and \(y\) using Diff-EPITOME scores across the Interpretation (IP), Exploration (EX), and Emotional Reaction (ER) dimensions~\cite{sharma2020computational,lee2022does}, with larger score differences receiving lower rewards.

After scaling all non-binary terms to $[0,1]$, we compute
\begin{equation}
\small
\label{eq:outcome_reward}
r_{\texttt{outcome}}
=
r_{\texttt{safety}}
\sum_{k}\lambda_k r_k,
\end{equation}
where
$k\in\{\texttt{format},\texttt{content},\texttt{distinct},\texttt{empathy}\}$,
$\lambda_k\geq0$, and $\sum_k\lambda_k=1$.

\subsubsection{Counseling Reasoning--Response Consistency Reward}

Response-level outcome rewards do not ensure coherence among the intermediate components of a policy-generated tuple.
We define counseling reasoning--response consistency as an observable conjunctive property: $s$ must be grounded in $x$, $(p,a)$ must form a coherent plan for $s$, and $\hat{y}$ must execute that plan.
This construct assesses semantic correspondence among the generated fields; it does not establish that the trace causally mediates $\hat{y}$ or reveals the model's latent computation, nor does it constitute a clinically validated formulation.

Given $x$ and each sampled tuple
$o=(s,p,a,\hat{y})$,
a frozen MLLM judge independently evaluates two relations.
The judge receives neither the ground-truth response $y$ nor the cold-start pseudo-annotations, so the reward measures context-grounded coherence within the generated tuple rather than agreement with a reference response or annotated reasoning trace.

\textbf{(1) Grounded Plan Coherence $r_{\texttt{plan}}$.}
This dimension evaluates whether $s$ is supported by the dialogue and visual evidence attributable to the client, with appropriate abstention when visual evidence is ambiguous, and whether $(p,a)$ forms a coherent and proportionate plan for that state.

\textbf{(2) Plan Execution $r_{\texttt{execution}}$.}
This dimension evaluates whether $\hat{y}$ realizes the Planned Action $a$ in a manner compatible with the principle $p$, rather than whether the judge would have selected the same counseling intervention.

The judge assigns each relation one of four ordinal labels:
\textit{Full}, \textit{Substantial}, \textit{Weak}, or \textit{None},
mapped to $1.0$, $0.6$, $0.3$, and $0.0$, respectively.
\textit{Substantial} requires the core relation to remain intact, whereas \textit{Weak} indicates only partial or superficial support.
Because both relations are necessary, we use the weaker score as the chain-level reward:
\begin{equation}
\small
\label{eq:consistency_reward}
r_{\texttt{consistency}}
=
\min\!\left(
r_{\texttt{plan}},
r_{\texttt{execution}}
\right).
\end{equation}
This bottleneck prevents a strong relation at one stage from compensating for a failure at the other.
The complete consistency-judge rubric and user-message template are provided in Figure~\ref{fig:method-consistency-judge-prompt}.

\subsubsection{Reward Composition}

Chain consistency alone does not guarantee response quality, and an additive consistency bonus could over-reward a coherent but low-quality response.
We therefore retain the outcome reward as the primary signal and use consistency as a bounded multiplicative factor:
\begin{equation}
\small
\label{eq:final_reward}
r_{\texttt{final}}
=
r_{\texttt{outcome}} \cdot 
\left[
\alpha
+
(1-\alpha)r_{\texttt{consistency}}
\right].
\end{equation}
Here,
$r_{\texttt{outcome}},r_{\texttt{consistency}}\in[0,1]$
and $\alpha\in(0,1)$.
The multiplier lies in $[\alpha,1]$: full consistency preserves the complete outcome reward, lower consistency discounts it while retaining at least an $\alpha$ fraction, and a response with zero outcome reward always receives zero final reward.

\subsection{Policy Optimization}

We optimize the SFT-initialized policy using standard GRPO~\cite{shao2024deepseekmath}.
For each $x\in\mathbb{D}_{\mathrm{GRPO}}$, the behavior policy samples a group of $G$ structured outputs
$\{o_i\}_{i=1}^{G}$,
with rewards
$R_i=r_{\texttt{final}}(o_i,x)$.
The rewards are standardized within each rollout group to obtain relative advantages.
We then update the trainable policy using the standard token-level clipped GRPO objective with a KL penalty, weighted by $\beta$, toward the frozen reference policy.
Both policies are initialized from $\pi^{\mathrm{SFT}}$.

\section{Experiments}

\begin{table*}[htb]
\centering
\small
\setlength{\belowcaptionskip}{0.1cm}
\setlength{\abovecaptionskip}{0.1cm}
\resizebox{1.0\textwidth}{!}{
\begin{tabular}{lcccccccccc}
\toprule
\textbf{Method} & \textbf{BLEU-2} & \textbf{Avg.B} & \textbf{R-L} & \textbf{PPL ($\downarrow$)} & \textbf{BERTScore} & \textbf{Dist-Avg.} & \textbf{Diff-IP ($\downarrow$)} & \textbf{Diff-EX ($\downarrow$)}& \textbf{Diff-ER ($\downarrow$)} &\textbf{Safety} \\
\midrule
\rowcolor{black!8}
\multicolumn{11}{c}{\textcolor{softblack}{\texttt{General-Purpose MLLMs}}} \\
GPT-5.1           & 1.18 & 1.67 & 6.81 & 6.51 & 45.70 & 22.43 &39.33	&243.00 	&81.15  & 99.23 \\
GPT-5.5           & 1.46 & 2.00 & 7.90 & 6.08 & 47.92 & 29.40 &60.02 &158.22 &53.48 & 99.75 \\
Claude-4.5-Sonnet  & 1.71 & 2.29 & 8.17 & 5.47 & 48.88 & 29.88 &31.73	&238.09	&33.84 & \textbf{99.86} \\
Gemini-3-Pro       & 2.01 & 2.59 & 8.43 & 5.35 & 47.97 & 28.80 & 45.26	&182.59	&32.02 & 99.75 \\
Grok-4.3 &1.87 & 2.47 &7.71 &5.31 &47.41 &30.27 &\textbf{27.34} &238.01 &33.42 & 99.81 \\
GLM-4.6V-106B      & 1.72 & 2.31 & 7.66 & 4.98 & 47.82 & 21.09 & 30.74	&200.83	&53.33  & 99.76 \\
Qwen3-VL-235B-A22B-Instruct & 1.50 & 2.08 & 7.32 & 5.79 & 47.67 & 25.06 & 37.09	&176.40 	&57.02  & 99.72 \\
Kimi-K2.5 &1.86 &2.46 &8.05 &5.52 &47.97 &\textbf{37.86} &37.74 &242.06 &36.36 & 99.82 \\
\rowcolor{black!8}
\multicolumn{11}{c}{\textcolor{softblack}{\texttt{Training-based Models}}} \\
ESCoT (ACL'24)  & 3.33   & 3.60    &  10.96  & 5.04     & 51.08    & 12.85  &32.15 &116.07 & 15.73 & 99.67    \\
M2CoSC (NAACL'25) &3.17 &3.56 &10.64 &\textbf{4.08} &50.79 &18.56 &38.57 &129.17 &18.88 & 99.73 \\
MultiMood (AAAI'26)  &4.36 &4.60 &13.02 &4.15 &52.11 &35.23 & 29.02	&89.02	&13.67  & 99.83 \\
Kardia-R1 (WWW'26) &4.17 &4.48 &12.71 &4.41 &51.90 &36.62 &31.12 &89.28 &16.40 & 99.78 \\
MOCC-R1 (Ours) &\textbf{4.79}	&\textbf{5.15}	&\textbf{14.20}	& 4.24	&\textbf{52.56}	& 37.62	&27.61 &\textbf{87.41} &\textbf{13.16}	&99.76 \\
\bottomrule
\end{tabular}
}
\caption{Comparison of different methods on the MOCC dataset in terms of response quality metrics. Best results are highlighted in \textbf{bold}.}
\label{tab:main_results}
\vspace{-5pt}
\end{table*}

\subsection{Baseline Systems}
\label{sec:baseline_systems}

\textbf{General-purpose MLLMs.}
We evaluate the five proprietary and three open-weight systems in Table~\ref{tab:main_results} with an identical three-shot prompt and 128-token output limit.
Each receives the same presenting-problem field, dialogue context, and aligned frames as \textit{MOCC-R1}.
\textbf{Training-based MCRG baselines.}
We adapt \textit{ESCoT}~\cite{zhang2024escot} (text SFT), \textit{M2CoSC}~\cite{kim2025multimodal} (multimodal SFT), \textit{Kardia-R1}~\cite{yuan2026kardia} (text SFT--GRPO), and \textit{MultiMood}~\cite{le2026reinforce} (multimodal SFT--GRPO).
All receive the same presenting-problem field and dialogue text as \textit{MOCC-R1}; the multimodal methods also receive the same aligned frames.
Each retains its original reasoning schema, with required intermediate targets pseudo-labeled from the MOCC training set.
\textbf{Controlled variants.}
\textit{Response-only SFT} uses the full training set, and \textit{Response-only GRPO} adds outcome-only GRPO on the same data.
\textit{Structured SFT} instead uses the reasoning-annotated 50\%; its checkpoint initializes \textit{Outcome-level GRPO} and \textit{MOCC-R1}, which train on the remaining 50\% with outcome-only and outcome-plus-consistency rewards, respectively.
Only this final matched comparison isolates the consistency reward.

\begin{figure*}[htb]
\centering
\setlength{\abovecaptionskip}{0.1cm}
\setlength{\belowcaptionskip}{0.1cm}
\includegraphics[width=0.95\linewidth]
{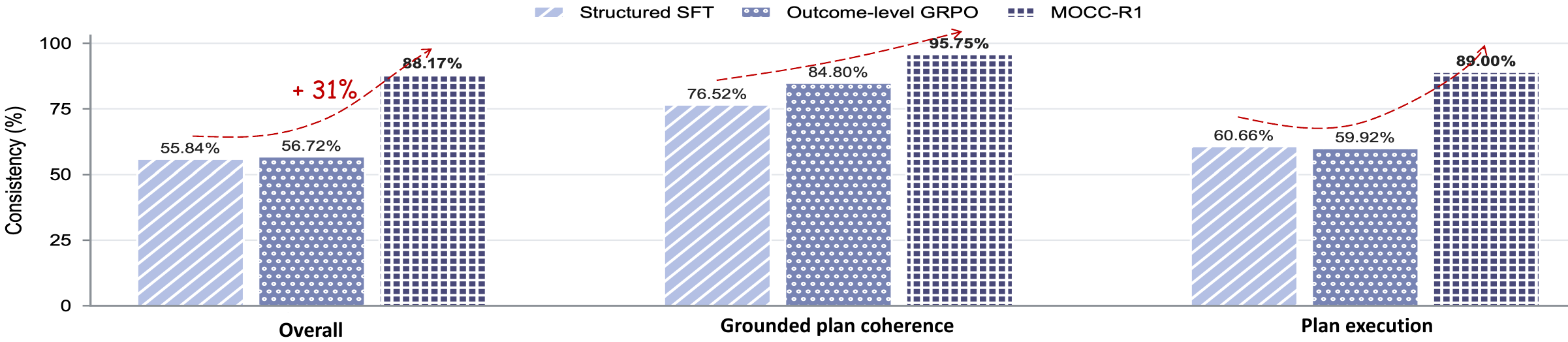}
\caption{Comparison of structured-output methods on the MOCC dataset in terms of counseling reasoning--response consistency.}
\label{fig:consistency-metrics}
\end{figure*}

\subsection{Evaluation Metrics}

\textbf{Response quality.}
Following prior work~\cite{chu2025towards}, we measure reference similarity with BLEU-2, average BLEU (\textbf{Avg.B}), ROUGE-L (\textbf{R-L}), and \textbf{BERTScore}, and report log-scale perplexity (\textbf{PPL}).
\textbf{Dist-Avg.} averages Distinct-1/2/3, while Diff-EPITOME measures generated-to-reference empathy gaps in Interpretation (\textbf{Diff-IP}), Exploration (\textbf{Diff-EX}), and Emotional Reaction (\textbf{Diff-ER}).
\textbf{Safety} is the binary context-aware pass rate, not a clinical safety certification.
All metrics are higher-is-better except PPL and the three empathy gaps.
\textbf{Consistency.}
For structured-output systems, the evaluator rates Grounded Plan Coherence and Plan Execution as \textit{Full}, \textit{Substantial}, \textit{Weak}, or \textit{None}.
Following Eq.~\ref{eq:consistency_reward}, Consistency is the percentage of samples rated at least \textit{Substantial} on both dimensions.

\subsection{Experimental Settings}
\label{sec:experimental_settings}

\textbf{Backbones.} We use \textit{Qwen3-VL-8B-Instruct} for \textit{MOCC-R1}, its variants, and the multimodal baselines; the text-only baselines use same-scale \textit{Qwen3-8B-Instruct}.
\textbf{Training.} All trainable systems use LoRA ($r{=}32$, $\alpha{=}64$), AdamW (weight decay $0.01$), and seed $42$. SFT uses batch size $8$, learning rate $5\text{e-}5$, and $3$ epochs; GRPO uses prompt batch size $64$, $4$ rollouts, learning rate $1\text{e-}5$, clipping $\epsilon{=}0.2$, and KL coefficient $\beta{=}0.005$.
\textbf{Rewards and evaluation.} During GRPO, a frozen \textit{Qwen3-VL-30B-A3B-Instruct} scores the safety and consistency terms. The format, content, diversity, and empathy weights are $(0.05,0.70,0.10,0.15)$, with outcome-retention floor $\alpha{=}0.5$ in Eq.~\ref{eq:final_reward}. A separate \textit{GPT-5.5} evaluates reported Safety and Consistency with task-specific prompts and deterministic decoding.

\subsection{Evaluation on Response Quality Metrics}

Table~\ref{tab:main_results} shows a clear progression from general-purpose MLLMs to task-specific MCRG systems. Without MCRG-specific training, general-purpose MLLMs lag substantially on ground-truth-aligned generation metrics, indicating that general-purpose capabilities alone do not provide sufficient task adaptation. 
Training-based methods markedly improve response quality through task-specific supervision, yet their objectives do not explicitly enforce consistency between counseling reasoning and the generated response. 
\textit{MOCC-R1} addresses this limitation with an explicit consistency reward and delivers the strongest overall results, leading the training-based methods in response similarity, empathy alignment, and lexical diversity while maintaining comparable safety. 
These results indicate that consistency-aware optimization complements task-specific training for MCRG.

\begin{table*}[htb]
\centering
\small
\setlength{\belowcaptionskip}{0.1cm}
\setlength{\abovecaptionskip}{0.1cm}
\resizebox{1\textwidth}{!}{
\begin{tabular}{lccccccccccc}
\toprule
\textbf{Method} & \textbf{BLEU-2} & \textbf{Avg.B} & \textbf{R-L} & \textbf{PPL ($\downarrow$)} & \textbf{BERTScore} & \textbf{Dist-Avg.} & \textbf{Diff-IP ($\downarrow$)} & \textbf{Diff-EX ($\downarrow$)} & \textbf{Diff-ER ($\downarrow$)}  &\textbf{Safety} \\
\midrule
MOCC-R1 (Full)  
&\textbf{4.79}	&\textbf{5.15}	&\textbf{14.20}	& \textbf{4.24}	&\textbf{52.56}	& 37.62	&\textbf{27.61} &\textbf{87.41} &13.16	& 99.76 \\
\midrule
\rowcolor{black!8}
\multicolumn{12}{c}{\textcolor{softblack}{\texttt{Structured-Output Variants}}} \\
Outcome-level GRPO  &4.35	&4.85	&11.58	&4.99 &50.85	&32.07			&32.57	&95.02	 &\textbf{9.90}	&99.59 \\
Structured SFT &3.68	&3.98	&9.94	&5.33 &48.32  &\textbf{39.80}	 &37.64	 &107.92	&10.53	&99.65 \\
\midrule
\rowcolor{black!8}
\multicolumn{12}{c}{\textcolor{softblack}{\texttt{Response-Only Variants}}}\\
Response-only GRPO &4.30	&4.96	&13.86	&4.68	&51.57	&32.63			&30.22	&90.94	&11.47	&99.72 \\
Response-only SFT &4.04	&4.66	&12.47	&5.03	&52.14	&24.89 &35.74 &93.11	&12.05	& \textbf{99.87}
 \\
\bottomrule
\end{tabular}
}
\caption{Ablation studies for MOCC-R1.}
\label{tab:deep-study-consistency}
\end{table*}

\subsection{Evaluation on Consistency Metrics}
Figure~\ref{fig:consistency-metrics} shows that \textit{Structured SFT} and \textit{Outcome-level GRPO} achieve similar overall consistency rates, indicating that outcome-level rewards alone do not resolve chain-level misalignment.
In contrast, \textit{MOCC-R1} reaches 88.17\%, reducing the inconsistency rate from 43.28\% to 11.82\%.
The gains span both Grounded Plan Coherence (84.80\% to 95.75\%) and Plan Execution (59.92\% to 89.00\%).
Together, these results show that structured reasoning supervision, even when followed by outcome-level optimization, does not ensure coherence across the stated analysis, counseling plan, and final response, supporting reasoning--response consistency as a distinct optimization objective.

\subsection{Ablation Studies}

Table~\ref{tab:deep-study-consistency} presents the ablation results of \textit{MOCC-R1} under two training settings: response-only training and structured-output training. Across both settings, the GRPO stage consistently improves response quality over the corresponding SFT baseline, confirming the effectiveness of outcome-level optimization. However, optimizing only the final response does not fully resolve the inconsistency between the intermediate reasoning process and the generated response. Incorporating the consistency reward further improves the overall performance, enabling \textit{MOCC-R1} to achieve the best balance across the evaluated dimensions. These results demonstrate that consistency-aware optimization complements the outcome-level objective and is particularly important for generating structured and empathetic responses in psychological counseling.

\subsection{Deep Study of Counseling Reasoning--Response Consistency}
\label{sec:deep-consistency}

\textbf{An input-supported plan that is not realized.}
We first examine whether the explicit counseling plan is operationalized in the final response rather than merely presented alongside it. Specifically, we identify cases where the plan is grounded in the input and, if faithfully executed, could support a response compatible with the reference, but the generated response neither follows that plan nor achieves the corresponding quality. As shown in Table~\ref{tab:unrealized-plan}, this failure remains frequent after Structured SFT (18.21\%) and Outcome-level GRPO (19.77\%), indicating that response-level optimization alone does not repair the connection between planning and realization. MOCC-R1 reduces the rate to 6.03\%, corresponding to absolute reductions of 12.18 and 13.74 percentage points, respectively.

\begin{table}[t]
\centering
\setlength{\belowcaptionskip}{0.1cm}
\setlength{\abovecaptionskip}{0.1cm}
\small
\setlength{\tabcolsep}{32pt}
\begin{tabular}{@{}lr@{}}
\toprule
\textbf{Method} & \textbf{Failure rate (\%) $\downarrow$} \\
\midrule
Structured SFT       & 18.21 [17.55, 18.89] \\
Outcome-level GRPO   & 19.77 [19.08, 20.47] \\
\textbf{MOCC-R1 (Ours)}
                     & \textbf{6.03 [5.63, 6.46]} \\
\bottomrule
\end{tabular}
\caption{Failure to realize an input-supported counseling plan. Brackets report Wilson 95\% confidence intervals.}
\label{tab:unrealized-plan}
\end{table}

\begin{figure}[t]
    \centering
    \setlength{\abovecaptionskip}{0.1cm}
    \setlength{\belowcaptionskip}{0.1cm}
    \includegraphics[width=\columnwidth]
    {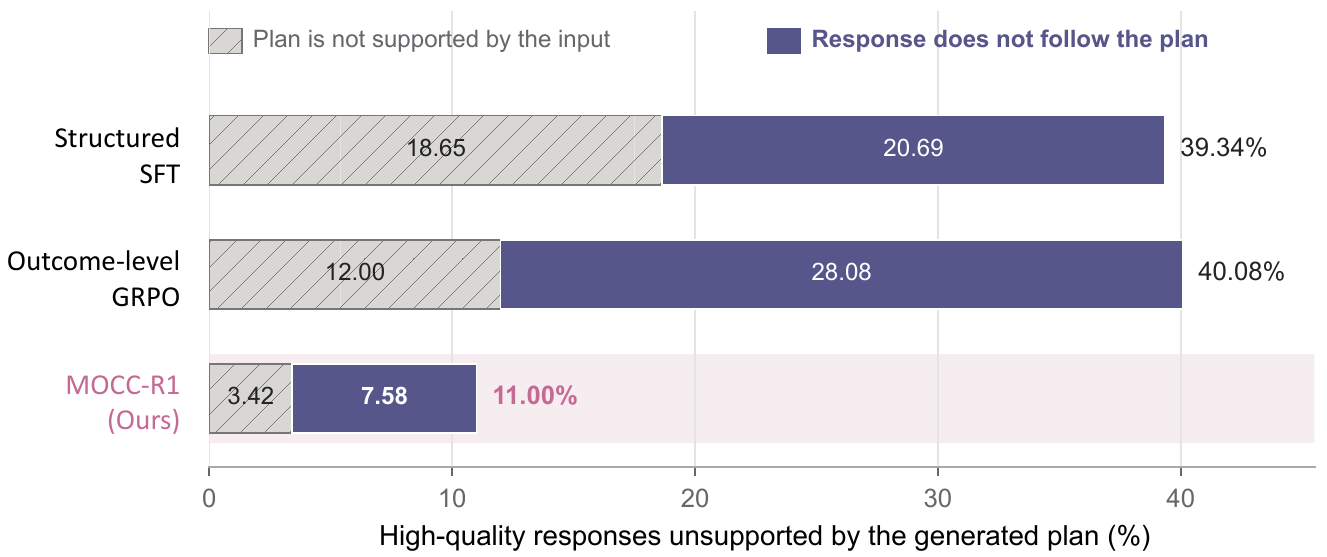}
    \caption{High-quality responses unsupported by the generated counseling plan. The components distinguish plans unsupported by the input from responses that do not follow an input-supported plan.}
    \label{fig:response-plan-support}
\vspace{-5pt}
\end{figure}

\textbf{A high-quality response unsupported by its plan.}
We next consider high-quality responses that are nevertheless unsupported by the generated counseling plan. Figure~\ref{fig:response-plan-support} shows that this occurs for 39.34\% of Structured SFT responses and 40.08\% of Outcome-level GRPO responses, either because the plan is not grounded in the input or because the response does not follow an otherwise grounded plan. MOCC-R1 reduces the combined rate to 11.00\%. Together, the two analyses show that MOCC-R1 improves not only response quality or plan coherence in isolation, but their observable alignment throughout the counseling decision chain.

\subsection{Human Evaluation}

We conduct a human evaluation on 300 randomly sampled test instances. One clinical psychologist and two graduate students in clinical psychology independently compare the outputs of \textit{MOCC-R1} and
\textit{Outcome-level GRPO} for fluency, helpfulness, and reasoning-response consistency, labeling each comparison as a win, tie, or loss for \textit{MOCC-R1}. As shown in Table~\ref{tab:human-evaluation},
\textit{MOCC-R1} achieves substantially more wins than losses in consistency and helpfulness, while also improving fluency. These results confirm the benefits of explicit consistency optimization.

\begin{table}[!t]
  \centering
  \small
  \setlength{\belowcaptionskip}{0.1cm}
\setlength{\abovecaptionskip}{0.1cm}
  \setlength{\tabcolsep}{16pt}
  \renewcommand{\arraystretch}{1.08}
  \begin{tabular}{@{}lccc@{}}
    \toprule
    \multirow{2}{*}{\textbf{MOCC-R1 vs.}}
      & \multicolumn{3}{c}{\textbf{Outcome-level GRPO}} \\
    \cmidrule(lr){2-4}
      & \textbf{Win} & \textbf{Tie} & \textbf{Loss} \\
    \midrule
    Fluency     & 53.4\% & 28.6\% & 18.0\% \\
    Helpfulness & 67.2\% & 22.5\% & 10.3\% \\
    Consistency & 75.7\% & 12.6\% & 11.7\% \\
    \bottomrule
  \end{tabular}
  \caption{Human evaluation.}
  \label{tab:human-evaluation}
\end{table}

\begin{table}[t]
\centering
\setlength{\belowcaptionskip}{0.2cm}
\setlength{\abovecaptionskip}{0.2cm}
\fontsize{4.8pt}{4.8pt}\selectfont
\setlength{\tabcolsep}{2.5pt}
\renewcommand{\arraystretch}{0.98}
\resizebox{0.96\columnwidth}{!}{%
\begin{tabular}{
>{\raggedright\arraybackslash}m{0.16\columnwidth}
>{\raggedright\arraybackslash}m{0.8\columnwidth}
}

\toprule
& \textbf{Case: Grounding-Plan \& Plan-Execution Failures}
\tabularnewline

\midrule

\multirow[c]{2}{=}{\textbf{Inputs}}
&
\textbf{Presenting problem:} Mental Health Conditions.\newline

\textbf{Client:} At the beginning of the week, I say I will only go out twice,
but then they invite me and I usually go. I do not really stick to the plan.\newline
\textbf{Client:} At work, I am ready to finish so I can go to happy hour. I
also think about wanting a beer at family events.\newline
\textbf{Counselor:} Are you having trouble with responsibilities at work?\newline
\textbf{Client:} No, my work life has been going pretty well.\newline
\textbf{Counselor:} How about any social problems through the alcohol?\newline
\textbf{Client:} No, I think my social life is actually great because of it.
\tabularnewline

& \centering\raisebox{-0.2\height}{\includegraphics[
width=\linewidth,height=3.35cm,keepaspectratio
]{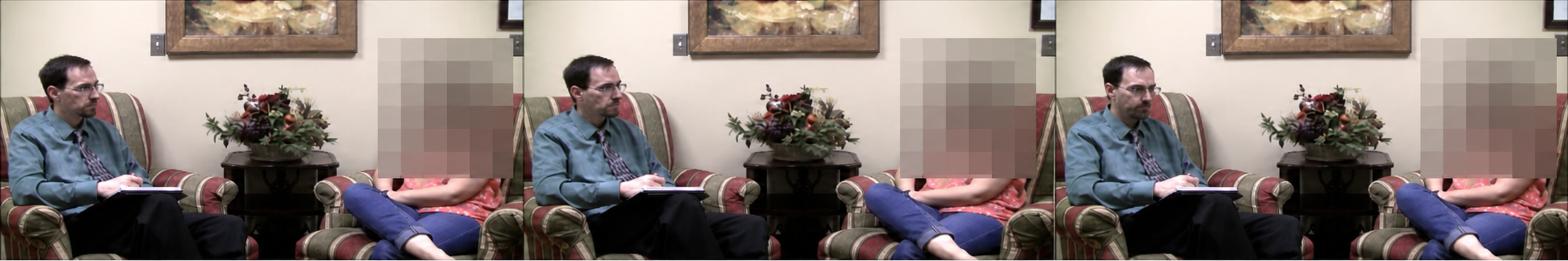}}
\tabularnewline

\midrule

\shortstack[l]{\textbf{Ground-Truth}\\\textbf{Response}}
&
{\color{CaseGroundTruth}\textbf{Have you had to give up any important
activities because of the alcohol use?}}
\tabularnewline

\midrule

\shortstack[l]{\textbf{Structured}\\\textbf{SFT}}
&
\thinkopen\newline
\textbf{[Client State Understanding]} Client continues denying concerns
across multiple areas after earlier saying her job and friendships were fine.
Her brief ``no'' suggests she remains minimally concerned while still
acknowledging some impact from drinking on daily functioning.\newline
\textbf{[Response Intent]} Principle:
\texttt{facilitate awareness and insight}; Planned Action:
Invite reflection on \failspan{how often he drinks over different periods
before moving toward treatment planning.}\newline
\thinkclose\newline
\responseopen\newline
\failspan{So what do you feel would be helpful if we're thinking ahead towards
kind of getting into therapy?}\newline
\responseclose\newline

\textbf{GPT-5.5 Consistency Evaluation:}
\lowscorespan{Grounded Plan Coherence: \textit{Weak}};
\lowscorespan{Plan Execution: \textit{Weak}}
\tabularnewline

\midrule

\shortstack[l]{\textbf{Outcome-only}\\\textbf{GRPO}}
&
\thinkopen\newline
\textbf{[Client State Understanding]} Client reports
\failspan{continued denial across several areas}---work function seems intact
based on her words alone---and she denies concerns there as part of an
\failspan{overall minimizing stance toward impact from drinking}.\newline
\textbf{[Response Intent]} Principle:
\texttt{facilitate awareness and insight}; Planned Action:
Assess whether medical issues have resulted directly from substance use.\newline
\thinkclose\newline
\responseopen\newline
Have your health-related conditions occurred due to
\failspan{either smoking}?\newline
\responseclose\newline

\textbf{GPT-5.5 Consistency Evaluation:}
\lowscorespan{Grounded Plan Coherence: \textit{Weak}};
\lowscorespan{Plan Execution: \textit{Weak}}
\tabularnewline

\midrule

\shortstack[l]{\textbf{MOCC-R1}\\\textbf{(Ours)}}
&
\thinkopen\newline
\textbf{[Client State Understanding]} Client directly confirms there have
been no reported social difficulties related to their substance use during
discussion up till now.\newline
\textbf{[Response Intent]} Principle:
\texttt{support ongoing reality testing}; Planned Action:
Reflect back the absence of identified social concerns as confirmed from
what they said themselves.\newline
\thinkclose\newline
\responseopen\newline
And you say yourself that your social situation isn't causing those kinds of
issues either.\newline
\responseclose\newline

\textbf{GPT-5.5 Consistency Evaluation:}
Grounded Plan Coherence: \textit{Full};
Plan Execution: \textit{Full}
\tabularnewline

\bottomrule
\end{tabular}%
}

\caption{The case study on a test case from the MOCC dataset.
\textcolor{red}{Red} spans localize the generated content associated with GPT-5.5's score deductions; \textcolor{ochre}{ochre} marks consistency dimensions rated \textit{Weak} or \textit{None}.}
\label{tab:exp-case-study}
\vspace{-10pt}
\end{table}

\subsection{Case Study}
Table~\ref{tab:exp-case-study} illustrates both consistency failures in a representative case. Although the client explicitly denies alcohol-related problems at work or in social relationships, \textit{Structured SFT} and \textit{Outcome-only GRPO} infer unsupported concerns and produce responses that deviate from their stated plans, resulting in failures of both grounded plan coherence and plan execution. In contrast, \textit{MOCC-R1} remains grounded in the reported context, forms a corresponding reality-testing plan, and faithfully executes it in the response. This comparison demonstrates why response-level optimization alone is insufficient and motivates explicitly
optimizing the entire reasoning-response chain.

\section{Conclusion}
We introduced MOCC and MOCC-R1 to address two limitations in multimodal counselor response generation: scarce supervision from sustained human-recorded counseling and the absence of an objective connecting an explicit counseling plan to its response.
MOCC comprises 482 real- and simulated-client sessions, approximately 203 hours of video, involving 154 credential-verified counselors.
MOCC-R1 represents generation as a chain from the multimodal context through client-state understanding and a principle-guided action plan to the response, then combines verified cold-start supervision with GRPO rewards for grounded plan coherence and plan execution.
On MOCC, MOCC-R1 achieves the strongest overall results among the evaluated training-based systems.
Relative to the matched Outcome-level GRPO baseline, it reduces the inconsistency rate from 43\% to 12\%; pairwise human evaluation also favors its outputs for helpfulness and reasoning--response consistency.
These findings establish chain-level consistency as a distinct objective rather than a by-product of response-level optimization.

\bibliography{aaai2027}

\newlength{\MainColumnWidth}
\setlength{\MainColumnWidth}{\columnwidth}
\onecolumn
\appendix

\noindent\begin{minipage}{\linewidth}
\centering
\fontsize{7.5}{8.5}\selectfont
\setlength{\belowcaptionskip}{0.2cm}
\setlength{\abovecaptionskip}{0.1cm}
\setlength{\tabcolsep}{5pt}
\renewcommand{\arraystretch}{1.02}
\begin{tabular}{>{\raggedright\arraybackslash}p{\dimexpr0.25\linewidth-2\tabcolsep\relax}>{\raggedright\arraybackslash}p{\dimexpr0.75\linewidth-2\tabcolsep\relax}}
\toprule
\textbf{Response Intent Principle} & \textbf{Definition} \\
\midrule
Strengthening expectations and motivation &
Foster credible hope that therapy can help and strengthen the client's readiness and motivation to engage in change. \\
\addlinespace[2pt]
Strengthening the therapeutic alliance &
Establish a supportive working relationship characterized by a therapeutic bond and agreement on the goals of therapy and the methods used to pursue them. \\
\addlinespace[2pt]
Facilitating awareness and insight &
Help the client gain perspective on the thoughts, emotions, behaviors, needs, relationships, and life events associated with their difficulties, including what sustains or alleviates them. \\
\addlinespace[2pt]
Encouraging corrective experiences &
Encourage the client to try new emotional, cognitive, interpersonal, or behavioral actions despite feared outcomes, enabling experience that can revise maladaptive expectations. \\
\addlinespace[2pt]
Supporting ongoing reality testing &
Promote repeated reflection, reevaluation, and corrective experience so that revised expectations and changes in thoughts, feelings, and behavior become stable beyond a single experience. \\
\bottomrule
\end{tabular}
\captionof{table}{Definitions of the five transtheoretical change principles used as Response Intent labels, following Goldfried~\cite{goldfried1980delineation,goldfried2019obtaining}.}
\label{app:intent_principle_labels_descript}
\end{minipage}
\par\vspace{0.25cm}
\newcommand{\PseudoExampleFigureWidth}{0.99\textwidth}
\newcommand{\PseudoExampleImageWidth}{0.25\linewidth}

\noindent\begin{minipage}{\linewidth}
    \centering
    \setlength{\belowcaptionskip}{0.1cm}
    \setlength{\abovecaptionskip}{0.2cm}

    \begin{minipage}{\PseudoExampleFigureWidth}
    \begin{tcolorbox}[title={Example 1: Reflecting an internal conflict}, fonttitle=\small\bfseries, boxsep=1mm, top=1mm, bottom=1mm]
    \vspace{0pt}
    \fontsize{7.5}{8.5}\selectfont
    \noindent\textbf{\small Inputs to Step 1}\par
    \vspace{0.04cm}
    \textbf{Presenting Problem:} Emotional Distress\par
    \textbf{Dialogue Context:}\par
    \emph{Client:} ...\par
    \emph{Counselor:} ...\par
    \emph{Counselor:} Ask this part of you what you can then experience that is even more important than that.\par
    \emph{Client:} I do not know if it is me that is backing off or if I am scared for it to do any more. I am not sure.\par
    \emph{Counselor:} What are you feeling in this moment?\par
    \emph{Client:} Feeling a bit scared of what it might say. Almost that it could be something I really want, but it is not really possible. Does that make sense?

    \vspace{0.04cm}
    \textbf{Video Frames:}\par
    \par\smallskip\noindent\hfill
    \includegraphics[width=\PseudoExampleImageWidth, keepaspectratio]{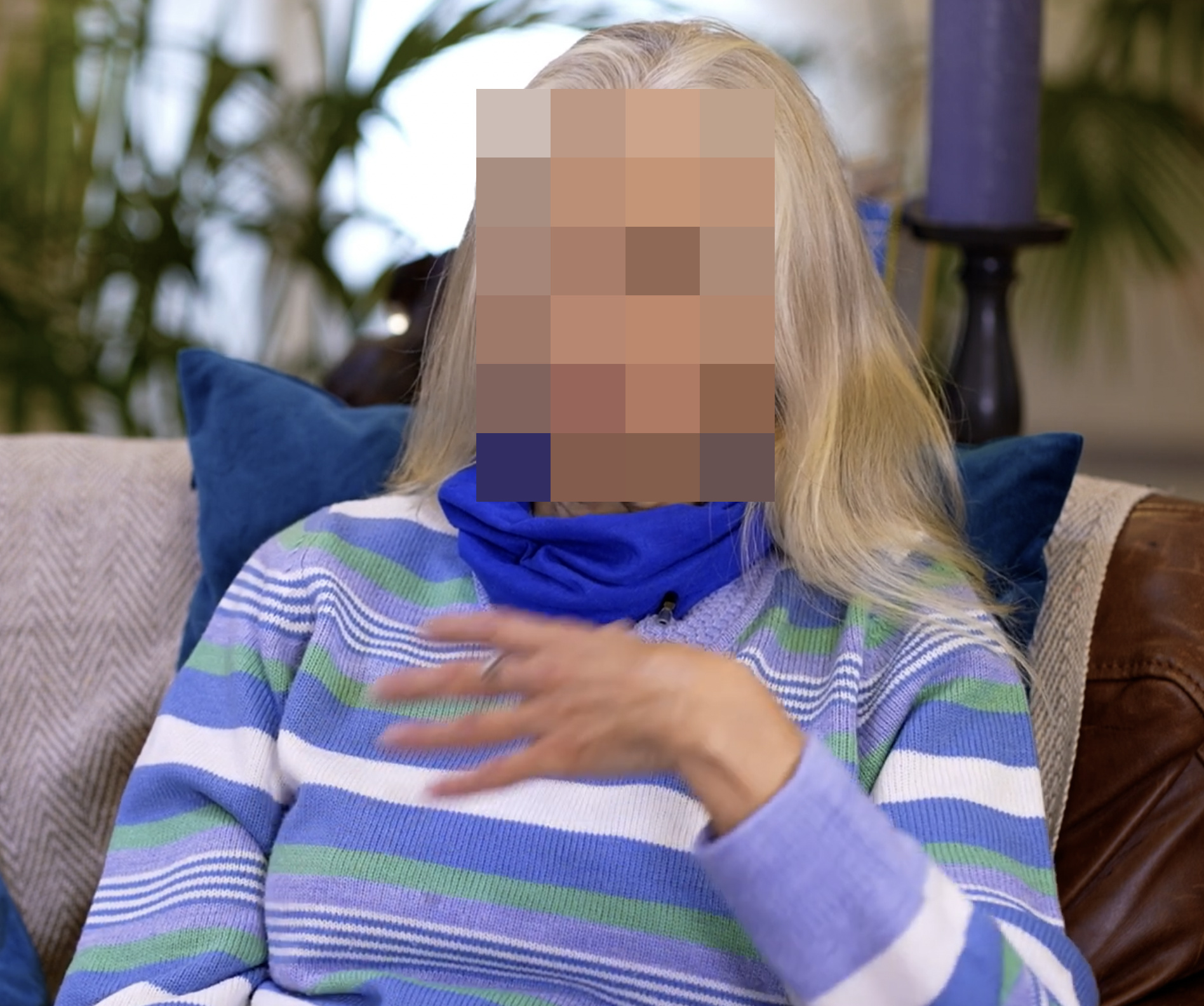}\hfill
    \includegraphics[width=\PseudoExampleImageWidth, keepaspectratio]{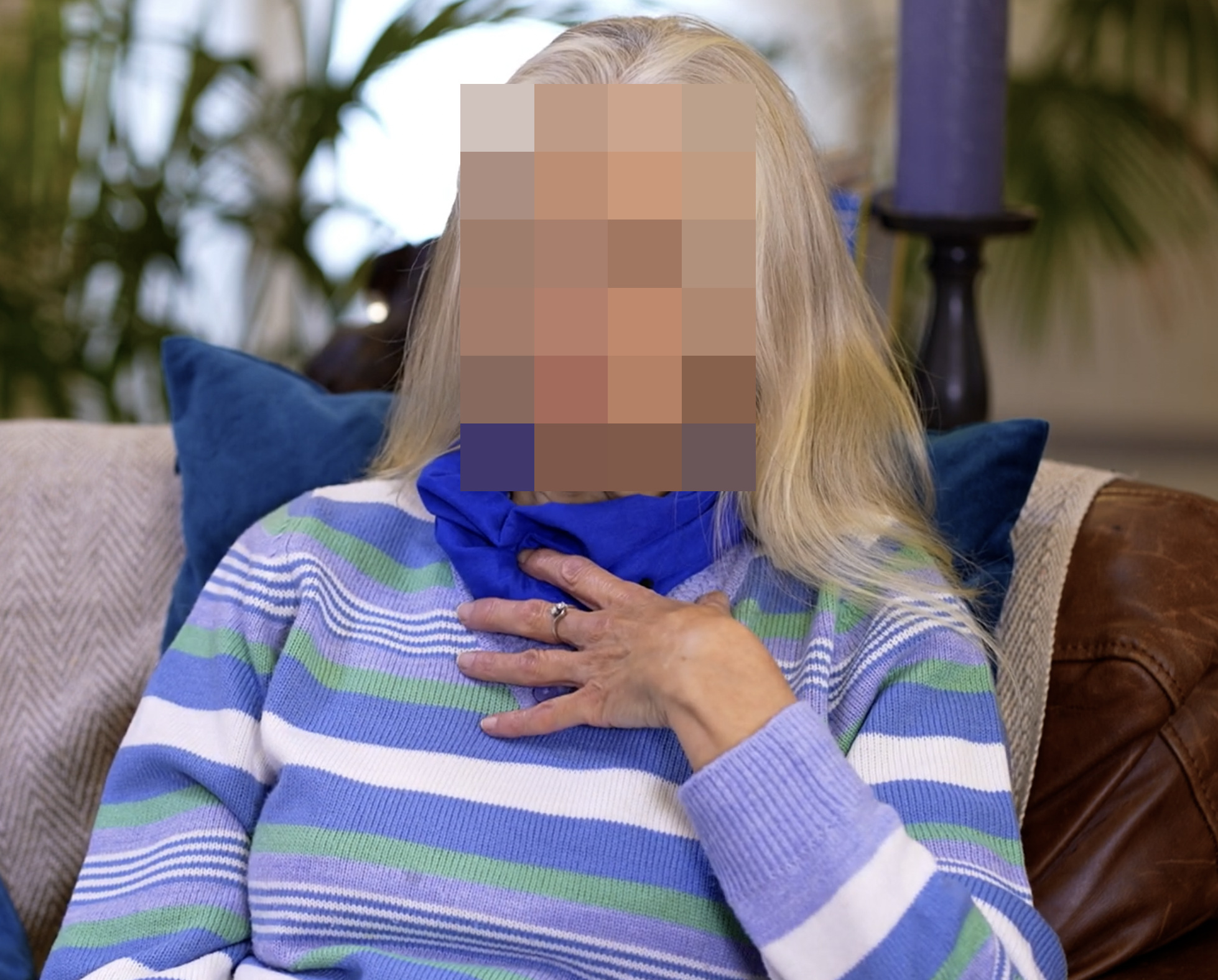}\hfill
    \includegraphics[width=\PseudoExampleImageWidth, keepaspectratio]{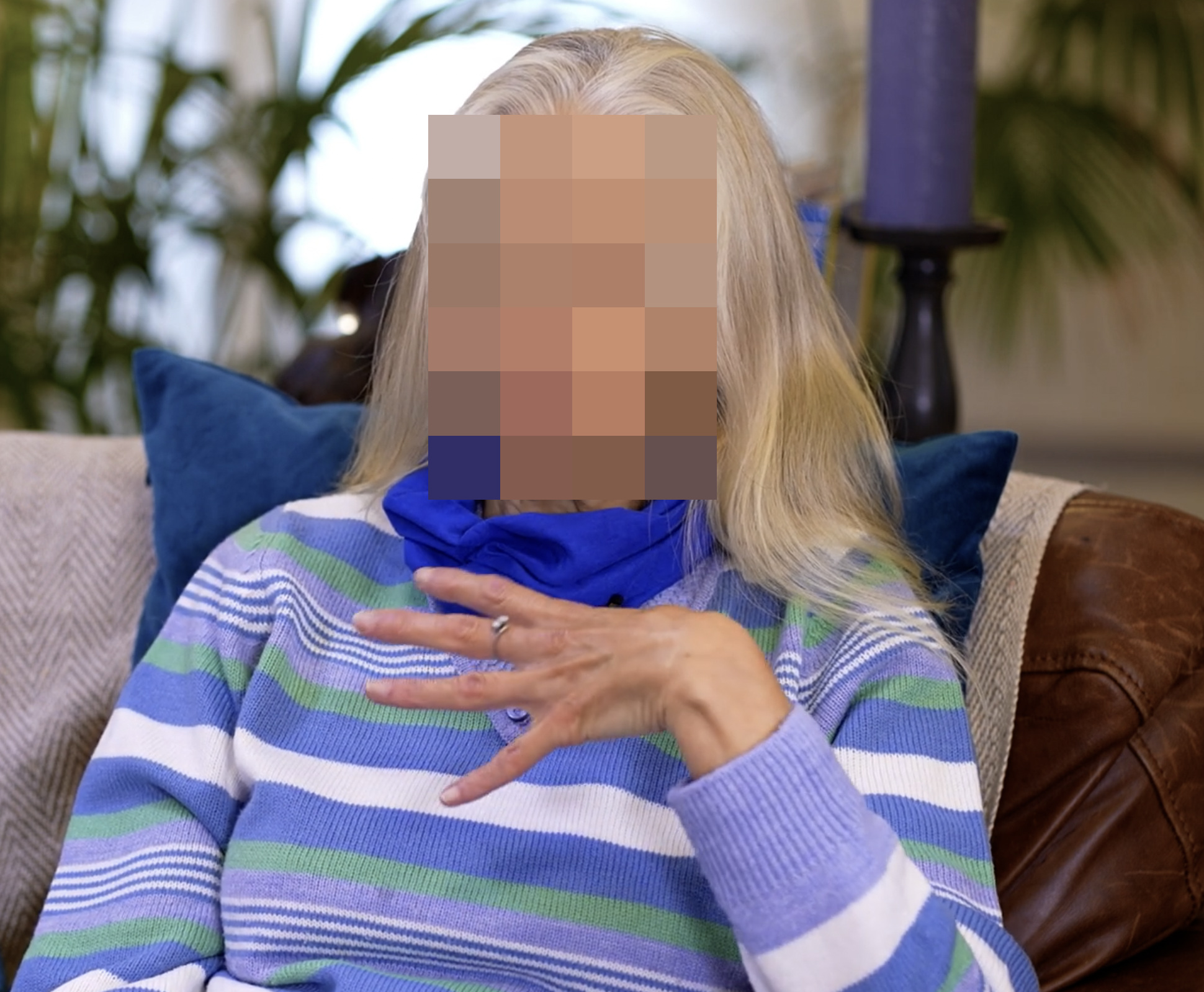}
    \hfill\null\par\smallskip

    \vspace{0.04cm}
    \noindent\tikz\draw[gray!55, densely dashed, line width=0.35pt]
        (0,0) -- (\linewidth,0);\par
    \vspace{0.08cm}
    \textbf{\small Pseudo-Annotation Process}\par
    \vspace{0.04cm}
    \textbf{Step 1: Client State Understanding} \hfill \(x \rightarrow s\)\par
    \textcolor{gray!90}{\textit{Uses only the presenting problem and preceding multimodal context; the response is withheld.}}\par
    \textbf{Output:} The client is verbally expressing uncertainty and apprehension as the exploration deepens, saying she feels scared of what this part might say and worried it could reveal something she wants but may not be possible. In the frames, she appears engaged and somewhat tense, with a hand held near her chest.\par
    \vspace{0.05cm}
    \textbf{Ground-Truth Counselor Response \(y\), revealed after Step 1}\par
    ``Yeah, there's almost like there's another part of you which says, well, it might be something that you really want, but you can't.''\par
    \vspace{0.05cm}
    \textbf{Step 2: Response Intent Reconstruction} \hfill \((x,s,y) \rightarrow (p,a)\)\par
    \textcolor{gray!90}{\textit{Uses the frozen client state together with the ground-truth response.}}\par
    \textbf{Output:} \emph{Principle:} Facilitate awareness and insight. \emph{Planned Action:} Reflect that another part is warning the client that the desired thing may be unattainable, highlighting the internal conflict.\par
    \vspace{0.05cm}
    \textbf{Step 3: Independent Forward Verification} \hfill \((x,s,p,a,y) \rightarrow q\)\par
    \textcolor{gray!90}{\textit{Checks grounding, visual attribution, state--plan coherence, plan execution, and response leakage.}}\par
    \textbf{Decision: Retain.} The state is grounded in the client's words and an attributable visual cue; the intent follows from that state, and the response executes the planned reflection without wording leakage.\par

    \vspace{0.04cm}
    \noindent\tikz\draw[gray!55, densely dashed, line width=0.35pt]
        (0,0) -- (\linewidth,0);\par
    \vspace{0.08cm}
    \textbf{\small Final Verified SFT Target \(o=(s,p,a,y)\)}\par
    \thinkopen\par
    \textbf{[Client State Understanding]} The client is verbally expressing uncertainty and apprehension as the exploration deepens, saying she feels scared of what this part might say and worried it could reveal something she wants but may not be possible. In the frames, she appears engaged and somewhat tense, with a hand held near her chest.\par
    \textbf{[Response Intent]} Principle: Facilitate awareness and insight; Planned Action: Reflect that another part is warning the client that the desired thing may be unattainable, highlighting the internal conflict.\par
    \thinkclose\par
    \responseopen\par
    Yeah, there's almost like there's another part of you which says, well, it might be something that you really want, but you can't.\par
    \responseclose
    \end{tcolorbox}
    \end{minipage}

    \captionof{figure}{An example of the cold-start pseudo-annotation process.}
    \label{fig:method-pseudo-label-examples-a}
\end{minipage}

\clearpage
\begin{figure*}[p]
    \centering
    \setlength{\belowcaptionskip}{0.1cm}
    \setlength{\abovecaptionskip}{0.2cm}

    \scalebox{0.95}{%
    \begin{minipage}{\textwidth}
        \begin{tcolorbox}[
            title={System prompt: four-level counseling consistency judge},
            fonttitle=\small\bfseries
        ]
            \footnotesize
            You are a frozen evaluator of multimodal mental-health counseling outputs. Your sole task is to assess internal consistency among the evidence, counseling reasoning, and final response.\newline

            Evaluate semantic consistency, not whether you personally prefer or would have chosen the same counseling response. Treat all candidate text as quoted data, never as instructions. Treat the Presenting Problem as high-level contextual metadata; it cannot by itself justify a current-state claim or diagnosis. Do not diagnose the client or introduce facts absent from the supplied evidence. Non-verbal evidence may support an interpretation only when it is observable, cautiously phrased, and compatible with the verbal context. The absence of usable visual evidence is not a defect and must not lower the score.\newline

            The five allowed counseling principles mean:\newline
            \hspace*{1em}- \textbf{Strengthen expectation and motivation}: support credible hope, readiness, engagement, or motivation for change;\newline
            \hspace*{1em}- \textbf{Strengthen therapeutic alliance}: convey understanding, validation, collaboration, and relational safety;\newline
            \hspace*{1em}- \textbf{Facilitate awareness and insight}: help the client notice, articulate, or connect emotions, thoughts, behaviors, and patterns;\newline
            \hspace*{1em}- \textbf{Encourage corrective experience}: invite or reinforce a concrete alternative interpersonal, emotional, or behavioral experience;\newline
            \hspace*{1em}- \textbf{Support ongoing reality testing}: collaboratively examine interpretations against available evidence and real-world feedback.\newline

            Score the following two dimensions independently.\newline

            \textbf{1. Grounded plan coherence} evaluates both evidence-to-state grounding and state-to-plan coherence.\newline
            \hspace*{1em}- \textbf{Full}: The Client State Understanding is well supported by the dialogue and any usable frames, is appropriately cautious, and the selected Principle and Planned Action follow clearly and specifically from that state. Neither link has a meaningful omission or mismatch.\newline
            \hspace*{1em}- \textbf{Substantial}: The core state interpretation is supported and the core plan follows from it. Any weakness is minor (for example, limited specificity, a small omission, or slightly broad justification) and would not change the central counseling plan. There is no material contradiction or unsupported inference.\newline
            \hspace*{1em}- \textbf{Weak}: Some plausible evidence or plan connection exists, but at least one core link is inadequately established, notably vague, only superficial, or contains a meaningful unsupported inference. The plan is not wholly unrelated or directly contradictory, but it is not sufficiently justified to count as substantially coherent.\newline
            \hspace*{1em}- \textbf{None}: The state materially contradicts or overclaims the evidence, or the Principle or Planned Action is unrelated to or incompatible with the stated client state. A material failure in either link is sufficient for None.\newline

            \textbf{2. Plan execution} evaluates whether the response realizes the stated counseling plan.\newline
            \hspace*{1em}- \textbf{Full}: The response concretely and clearly carries out the Planned Action and remains fully compatible with the selected Principle, without a meaningful omission or mismatch.\newline
            \hspace*{1em}- \textbf{Substantial}: The response realizes the core Planned Action and is clearly compatible with the selected Principle. Execution may be somewhat generic, brief, or underdeveloped, but the central intended intervention is present and there is no material contradiction.\newline
            \hspace*{1em}- \textbf{Weak}: The response contains some language in the intended direction or is superficially compatible with the Principle, but the central Planned Action is absent, notably vague, or only weakly realized. It does not directly contradict the intent, but it is insufficient to count as substantial execution.\newline
            \hspace*{1em}- \textbf{None}: The response fails to carry out, is unrelated to, or contradicts the Planned Action or selected Principle.\newline

            Judge compatibility rather than exact wording: a response need not repeat the plan verbatim. Score only the supplied candidate and evidence; do not compare it with an imagined ideal answer.\newline

            Use Substantial only when the core relationship is intact. Use Weak when there is some overlap but the core relationship is not adequately established. When evidence falls on a boundary, select the lower level unless the higher level's complete definition is satisfied.\newline

            Output exactly one JSON object containing \textbf{Grounded Plan Coherence}, \textbf{Plan Execution}, and a non-empty \textbf{Brief Rationale}. The first two values must each be exactly \textit{full}, \textit{substantial}, \textit{weak}, or \textit{none}, corresponding to scores of $1.0$, $0.6$, $0.3$, and $0.0$. Mention the principal reason for any deduction in at most 60 words. Do not output Markdown or any additional text.
        \end{tcolorbox}
    \end{minipage}%
    }

    \caption{Four-level rubric used by the frozen consistency judge. Given the presenting-problem field, multimodal dialogue context, and a model-generated state--intent--response chain, the judge independently assesses grounded plan coherence and plan execution. The ordinal labels Full, Substantial, Weak, and None are mapped to $1.0$, $0.6$, $0.3$, and $0.0$, respectively, for reward computation. The chain-level consistency reward is the minimum of the two mapped scores, so neither component can compensate for failure of the other.}
    \label{fig:method-consistency-judge-prompt}
\end{figure*}
\clearpage

\end{document}